\documentclass[letterpaper, 10 pt, conference]{ieeeconf}  

\IEEEoverridecommandlockouts                              

\usepackage{graphics} 
\usepackage{epsfig} 
\usepackage{mathptmx} 
\usepackage{times} 
\usepackage{amsmath} 
\usepackage{amssymb}  
\usepackage{booktabs}
\usepackage{url}
\usepackage{multirow}
\usepackage{bbding}

\title{Achelous: A Fast Unified Water-surface Panoptic Perception Framework based on Fusion of Monocular Camera and 4D mmWave Radar}

\author{Runwei Guan$^{1\star}$, Shanliang Yao$^{1\star}$, Xiaohui Zhu$^{2}$, Ka Lok Man$^{2}$, Eng Gee Lim$^{2}$, Jeremy Smith$^{1}$,  Yong Yue$^{2}$, \\ Yutao Yue$^{3*}$
\thanks{$^{\diamond}$The authors acknowledge XJTLU-JITRI Academy of Industrial Technology for giving valuable support to the joint project. This work is supported by the Key Program Special Fund of XJTLU (KSF-A-19), Research Development Fund of XJTLU (RDF-19-02-23). Suzhou Science and Technology Fund (SYG202122). This work received financial support from Jiangsu Industrial Technology Research Institute (JITRI) and Wuxi National Hi-Tech District (WND).}
\thanks{$^{\star}$Runwei Guan and Shanliang Yao contribute equally.}
\thanks{$^{1}$Runwei Guan, Shanliang Yao and Jeremy Smith are with Faculty of Science and Engineering,
        University of Liverpool, L69 3BX Liverpool, United Kingdom. 
        {\tt\small \{runwei.guan, shanliang.yao, J.S.Smith\}@liverpool.ac.uk}}%
\thanks{$^{2}$Xiaohui Zhu, Ka Lok Man, Eng Gee Lim and Yong Yue are with School of Advanced Technology, Xi'an Jiaotong-Liverpool University, 215123 Suzhou, China.  
        {\tt\small \{xiaohui.zhu, Ka.Man, enggee.lim, yong.yue\}@xjtlu.edu.cn}}%
\thanks{$^{3*}$Yutao Yue, corresponding author, is with Institute of Deep Perception Technology, JITRI, 214000 Wuxi, China.  
        {\tt\small yueyutao@idpt.org}}%
}

\begin{document}

\maketitle
\thispagestyle{empty}
\pagestyle{empty}

\begin{abstract}
Current perception models for different tasks usually exist in modular forms on Unmanned Surface Vehicles (USVs), which infer extremely slowly in parallel on edge devices, causing the asynchrony between perception results and USV position, and leading to error decisions of autonomous navigation. Compared with Unmanned Ground Vehicles (UGVs), the robust perception of USVs develops relatively slowly. Moreover, most current multi-task perception models are huge in parameters, slow in inference and not scalable. Oriented on this, we propose Achelous, a low-cost and fast unified panoptic perception framework for water-surface perception based on the fusion of a monocular camera and 4D mmWave radar. Achelous can simultaneously perform five tasks, detection and segmentation of visual targets, drivable-area segmentation, waterline segmentation and radar point cloud segmentation. Besides, models in Achelous family, with less than around 5 million parameters, achieve about 18 FPS on an NVIDIA Jetson AGX Xavier, 11 FPS faster than HybridNets, and exceed YOLOX-Tiny and Segformer-B0 on our collected dataset about 5 mAP$_{\text{50-95}}$ and 0.7 mIoU, especially under situations of adverse weather, dark environments and camera failure. To our knowledge, Achelous is the first comprehensive panoptic perception framework combining vision-level and point-cloud-level tasks for water-surface perception. To promote the development of the intelligent transportation community, we release our codes in \url{https://github.com/GuanRunwei/Achelous}.
\end{abstract}


\section{Introduction}

With the rapid development of deep learning, Autonomous Driving (AD) has been becoming highly intelligent. Perception, as an essential module of AD, assembles many environment perception tasks, including object detection, drivable-area segmentation, lane line segmentation, etc. Correspondingly, multi-sensor fusion has been applied to improve the precision and robustness of perception systems \cite{bai2022transfusion}.

\begin{figure}
    \centering
    \includegraphics[width=0.98\linewidth]{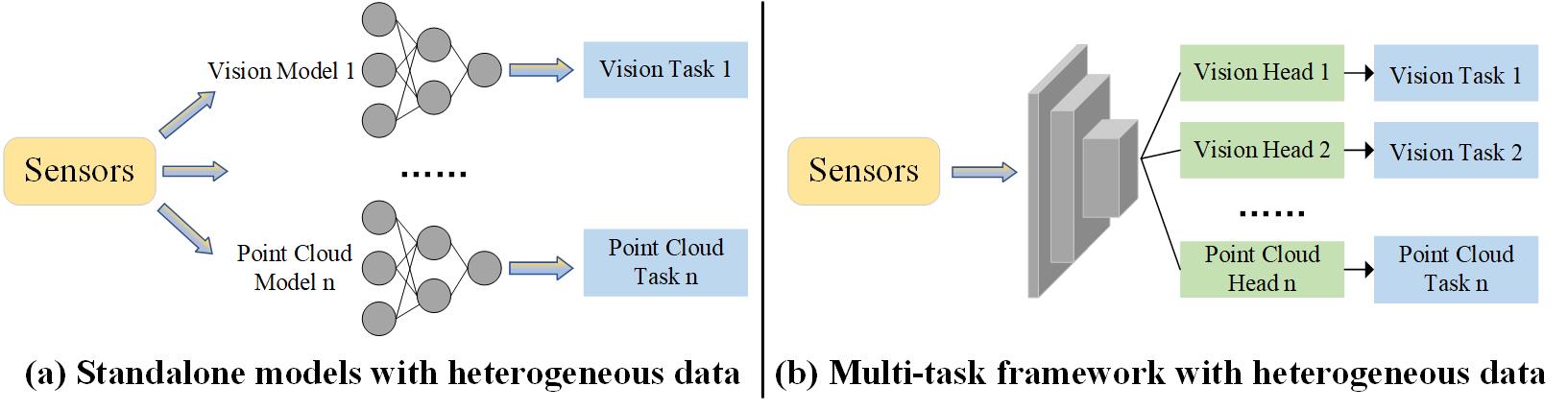}
    \caption{Comparison of (a) standalone models and (b) multi-task framework with heterogeneous sensor data. Our Achelous belongs to the latter.}
    \label{fig:overview}
\end{figure}

We have witnessed a remarkable development of unmanned ground vehicles (UGVs) with the help of deep learning. However, water-surface perception develops considerably slowly compared with road perception. Unmanned surface vehicles (USVs) play a significant role in water handling, such as water rescue \cite{yang2020maritime}, water quality monitoring \cite{zhu2019designing}, garbage collection \cite{madeo2020low} and geological exploration \cite{xue2019development}. Based on extensive surveys, firstly, we find most models of water-surface perception can only execute a single task \cite{haghbayan2018efficient}\cite{cheng2021robust}\cite{kim2021robust}, which is far to help USVs complete autonomous navigation. In addition, some researches show that multi-tasks can improve each other \cite{hybridnets}\cite{kendall2018multi}. Secondly, the industry prefers to parallelize multiple single-task models, which may lead to asynchronous perception results. Besides, perception system speed depends on the slowest model (Fig. \ref{fig:overview}). Thirdly, many models have never considered the problem of real-time inference on edge devices. They can only run on high-performance GPU devices \cite{hybridnets}\cite{yolop} at remote servers, which excessively relies on network communication capabilities. However, the network communication is weaken dramatically during ocean voyages, once the network is interrupted, it may be a catastrophe for USVs during some dangerous tasks. Last but not least, vision-only models \cite{zhang2019real}\cite{liu2021sea} are unreliable when confronted with dark environments, dense fog or lens failure. Currently, 4D mmWave radar is considered a promising complementary perception sensor for cameras in adverse situations, but how to efficiently extract irregular radar features is a challenge. Based on the above,  

\begin{figure*}
    \centering
    \includegraphics[width=0.95\linewidth]{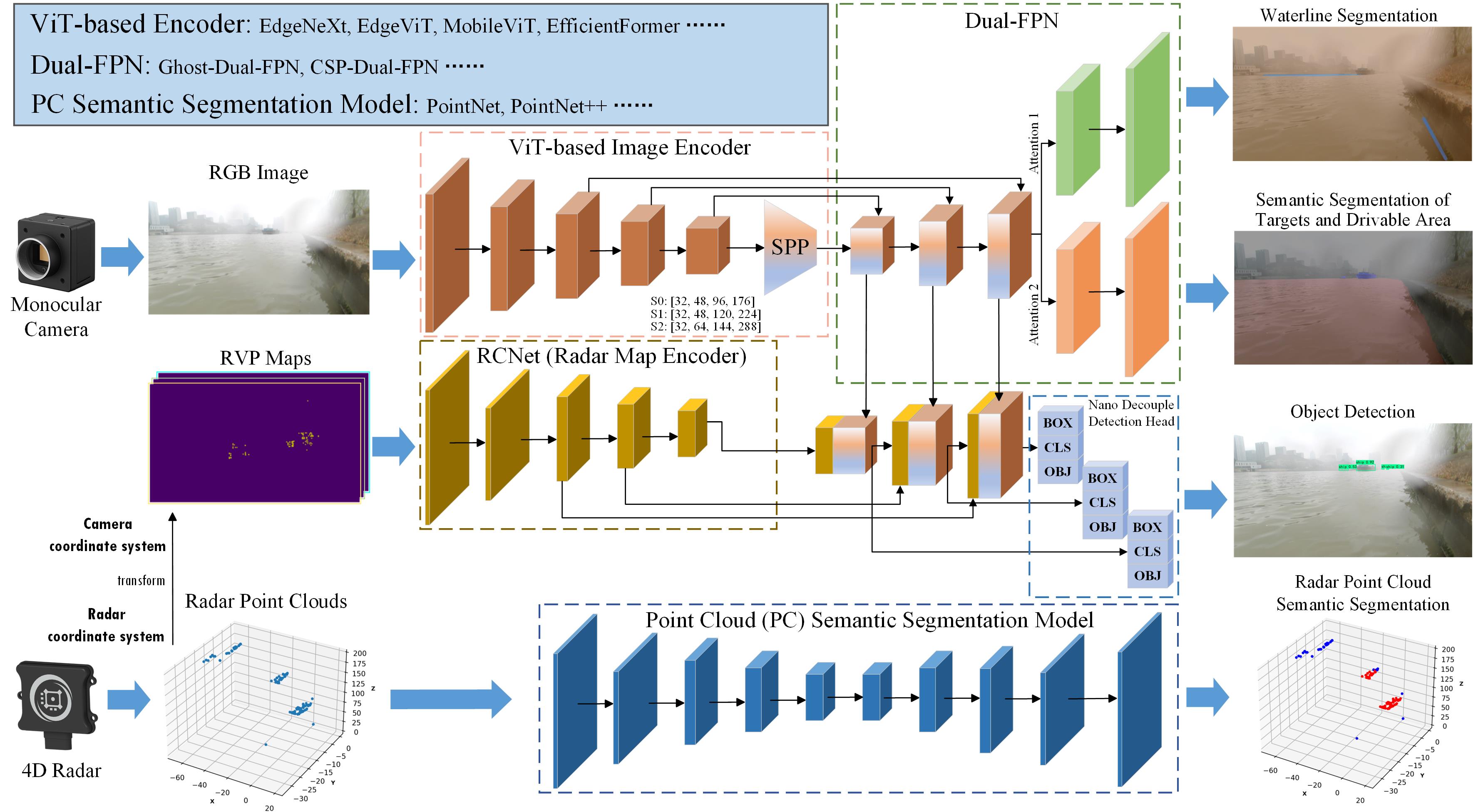}
    \caption{The architecture of Achelous. Blue point clouds in the semantic segmentation of radar point clouds denote clutter while the red one denotes target.}
    \label{fig:Achelous}
\end{figure*}

\begin{enumerate}
    \item We propose Achelous (Fig. \ref{fig:Achelous}), a low-cost and fast unified water-surface panoptic perception framework based on the fusion of a monocular camera and 4D radar. Achelous ensembles five perception tasks in one end-to-end framework, including object detection, object semantic segmentation, waterline segmentation, drivable-area segmentation and radar point cloud semantic segmentation. Achelous obtains about 18 FPS on an NVIDIA Jetson AGX Xavier and achieves competitive performances compared with single-task models and other multi-task models on perception tasks.
    \item We propose a simple but effective convolution operator called Radar Convolution (RadarConv). RadarConv is friendly to the irregularness of radar point clouds and can meticulously and effectively extract point cloud 
    features in 2D image planes, compared with the normal convolution.
    \item To promote the development of water-surface panoptic perception based on multiple sensors, our Achelous family are scalable and open-source.
\end{enumerate}

\section{Achelous}

\begin{table}
\setlength\tabcolsep{3.5pt}
\caption{Comparison of Panoptic Perception Models (Frameworks)}
\centering
\label{tab:pp_compare}
\begin{tabular}{cccccc}  
\toprule   
  Name & Type$^1$ & Sensor(s) & Tasks & Edge Test & Scalable \\
\midrule 
  YOLOP \cite{yolop} & M & camera & 3 & \XSolidBrush & \XSolidBrush   \\
  HybridNets \cite{hybridnets} & M & camera & 3 & \XSolidBrush & \XSolidBrush    \\
\midrule 
  \multirow{2}*{\textbf{Achelous (ours)}} & \multirow{2}*{\textbf{F}} & \textbf{camera} & \multirow{2}*{\textbf{5}} & \multirow{2}*{\CheckmarkBold} & \multirow{2}*{\CheckmarkBold} \\
   & & \textbf{4D radar} & & \\
\bottomrule  
\end{tabular}\\
\vspace{1mm}
\footnotesize{1. \textbf{M}odel or \textbf{F}ramework.}
\end{table}

\subsection{Overview}

As Fig. \ref{fig:Achelous} presents, our Achelous is based on a USV mounted with a monocular camera and 4D radar. The monocular camera captures RGB images while 4D radar obtains 3D point clouds directly. Each radar point cloud contains several physical features of targets. Among these physical features, we select the range, velocity and reflected power of the target, which cannot be perceived by the camera. To make radar point clouds assist vision-based object detection, we transform the coordinates of point clouds from the 3D radar coordinate system to 2D camera plane. We call 2D radar pseudo images RVP maps, where each channel represents the radar target's range, velocity and power. 

The main body of our Achelous contains four parts, a ViT-based image encoder, a radar feature encoder, prediction heads and a point cloud semantic segmentation model. As Table \ref{tab:pp_compare} presents, our Achelous supports more sensors and tasks than the other two panoptic perception models, YOLOP and HybridNets. Besides, Achelous specifically performs edge test, where the modules are optional and scalable. Achelous has three channel sizes, S0, S1 and S2. To accelerate inference, Achelous compresses branches and network fragmentation as much as possible, and weighs uses of activation function and group convolution. Besides, Achelous keep input and output channels of network unit equal to reduce memory access cost.

\subsection{ViT-based Image Encoder}
We have witnessed excellent performances of vision-transformer-based (ViT-based) models over the past years. ViT-based models can model global contextual features based on the self-attention mechanism \cite{dosovitskiyimage}. In addition, ViTs overall exceed CNNs in predictive performances \cite{liu2021swin}\cite{liu2022swin}\cite{baobeit}. ViTs are more robust than CNNs on adversarial attacks \cite{Shao_Shi_Yi_Chen_Hsieh_2021}\cite{Bhojanapalli_Chakrabarti_Glasner_Li_Unterthiner_Veit_2021}\cite{Paul_Chen_2022}, vision object occlusions \cite{Naseer_Ranasinghe_Khan_Hayat_Khan_Yang_2021} and data corruptions \cite{minderer2021revisiting}. Multi-head self-attention can assemble prediction features but CNNs cannot \cite{park2022blurs}, whose information capacity is much more than CNNs with the same parameters. Although ViTs are blamed for slow inference, recent studies \cite{pan2022edgevits}\cite{maaz2023edgenext}\cite{mehta2021mobilevit}\cite{li2022efficientformer} indicate that ViTs with ingenious design can still run as fast as CNNs. Based on the above advantages, our Achelous leverages ViT-based models as image encoders. Followed by the consistent paradigm of backbones, our image encoder has five stages with feature maps of multi-scale sizes, where the last four stages contain 2, 2, 6 and 4 layers, respectively. Our Achelous preliminarily contains four lightweight ViT-based backbones, EdgeNeXt \cite{maaz2023edgenext}, EdgeViT \cite{pan2022edgevits}, MobileViT \cite{mehta2021mobilevit} and EfficientFormer \cite{li2022efficientformer}. Following the backbone, spatial pyramid pooling (SPP) \cite{he2015spatial} is to enlarge receptive fields of image feature maps with multiple scales. 

\subsection{Dual-FPN and Segmentation Heads}
Feature Pyramid Network (FPN), as a significant module, is to fuse multi-scale features. Achelous has a dual FPN, fusing features extracted by the ViT-based encoder, where the first three feature maps are weight-shared and the last feature map is weight-independent. Between the weight-shared and weight-independent feature maps, we use two shuffle attention \cite{Zhang_Yang_2021} modules to remeasure features of two different segmentation tasks. Inspired by GhostNet \cite{Han_Wang_Tian_Guo_Xu_Xu_2020} and CSPDarknet \cite{Bochkovskiy_Wang_Liao_2020}, we design two lightweight FPNs, Ghost-Dual-FPN (GDF) and CSP-Dual-FPN (CDF), where GDF could dramatically remove feature redundancy in feature fusion stage while CDF could speed up feature fusion operations. Moreover, there are two segmentation heads following fused feature maps, which are for waterline segmentation, and semantic segmentation of targets and drivable-area. 

\subsection{Point Cloud Semantic Segmentation Model}
Since cameras may fail due to awful weather, radar would take over the perception of Achlous. As radar could not adopt the vision-based detection pattern, semantic segmentation of radar point clouds matters. Based on the ideas of permutation invariance and local feature learning from PointNet \cite{Charles_Su_Kaichun_Guibas_2017} and PointNet++ \cite{Qi_Yi_Su_Guibas_2017}, there are many models of point cloud processing. Here, to reduce the computation burden of the framework and make it fast, we adopt PointNet and PointNet++ as components of point cloud semantic segmentation. In addition, we reduce the number of channels of models to one-third of the original, because radar point clouds are dramatical sparser than lidars and too many latent channels are redundant.

\subsection{Radar Convolution, RCBlock, RCNet and Fusion with Image Features}
\label{subsec:rcnet}

\begin{figure}
    \centering
    \includegraphics[width=0.68\linewidth]{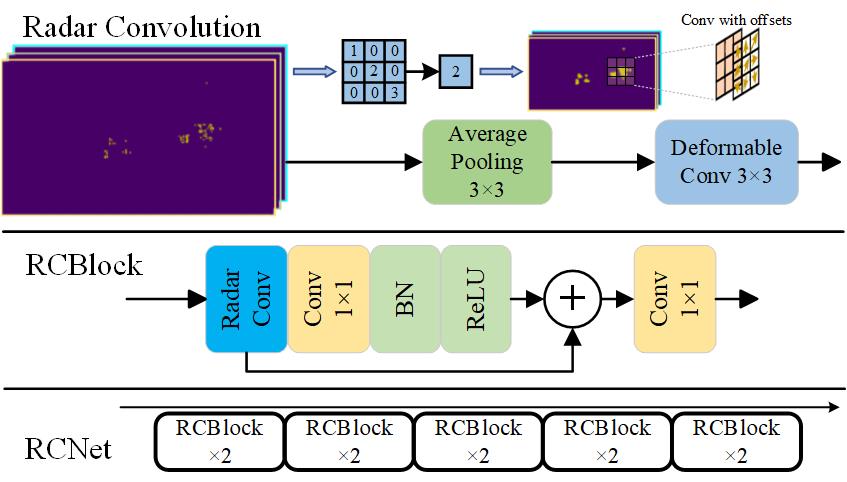}
    \caption{Radar Convolution, RCBlock and RCNet.}
    \label{fig:radarConv}
\end{figure}

We notice that radar point clouds are sparse and irregular, which means the conventional convolution contains many invalid operations and takes feature maps as regular grids. To make convolution friendly to feature extraction of radar point clouds and alleviate feature loss, we propose radar convolution (Fig. \ref{fig:radarConv}), a simple but highly-efficiency convolution operator. We first adopt average pooling with $3 \times 3$ to enlarge the receptive field. Compared to max-pooling, average pooling can keep more feature information, aggregating local features. Besides, the pooling operation is much faster than convolution. To model the irregularity of radar point clouds, we introduce the deformable convolution \cite{zhu2019deformable}, which extract feature with offsets. Based on radar convolution, we construct RCBlock and RCNet as shown in Fig. \ref{fig:radarConv}. RCBlock contains two $1 \times 1$ convolutions to weigh each spatial feature. The number of channels in RCNet is one-quarter of that in ViT-based image encoder, since radar point clouds are sparse and do not need so many latent features with non-linear operations. 

RCNet is an auxiliary network for detection, where radar feature maps are concatenated to image feature maps in the dual-FPN, to help Achelous localize targets faster and improve the recall under adverse situations. Although many works fuse radar and image features in both backbone and neck \cite{cheng2021robust}\cite{song2022ms}, we find too many branches for fusion will cause a dramatic drop in inference speed. Since image feature maps in the dual-FPN contain abundant detailed low-level features for segmentation, in addition, upsampling operations and SPP equip feature maps with multi-scale features in different stages. Therefore, FPN-stage fusion is enough for robust object detection.

\subsection{Nano Decouple Detection Head}
We decouple feature maps in the detection head to predict bounding boxes, categories and confidence, respectively. In addition, we adopt depth-wise separable convolution to reduce parameters to a great extent. Furthermore, anchor-free is to accelerate inference and SimOTA \cite{ge2021yolox} algorithm is to improve matching positive samples.

\section{Experiments}

\begin{table*}
\center
\footnotesize
\setlength\tabcolsep{2.2pt}
\caption{Performances of Achelous, Other Multi-task Models and Single-task Models on Our Testset.}
  \label{tab:prediction_results}
\begin{tabular}{l|cc|cc|ccc|cc|c|c|cc}
\toprule
\multicolumn{1}{c}{\multirow{2}[2]{*}{\textbf{Methods}}} &
\multicolumn{1}{c}{\multirow{2}[2]{*}{\textbf{Sensors}}} &
\multicolumn{1}{c}{\multirow{2}[2]{*}{\textbf{TN$^1$}}} & 
\multicolumn{1}{c}{\multirow{2}[2]{*}{\textbf{Params (M)}}} & 
\multicolumn{1}{c}{\multirow{2}[2]{*}{\textbf{FLOPs (G)}}} & 
\multicolumn{3}{c}{\bf{OD$^2$}} &
\multicolumn{2}{c}{\bf{SS$^3$}} & 
\multicolumn{1}{c}{\bf{WS$^4$}} & 
\multicolumn{1}{c}{\bf{PC-SS$^5$}} & 
\multicolumn{1}{c}{\multirow{2}[2]{*}{\textbf{FPS$_{\text{e}}^{14}$}}} &
\multicolumn{1}{c}{\multirow{2}[2]{*}{\textbf{FPS$_{\text{g}}^{15}$}}}
  \\ \cmidrule(lr){6-8}\cmidrule(lr){9-10}\cmidrule(lr){11-11}\cmidrule(lr){12-12}
\multicolumn{5}{c}{} & \bf{mAP$_{50\text{-}95}$} & \bf{mAP$_{50}$} & \bf{AR$_{50\text{-}95}$} & \bf{mIoU$_{\text{t}}^{12}$} & \bf{mIoU$_{\text{d}}^{13}$} & \bf{mIoU} & \bf{mIoU}     
\\\midrule
\textbf{Achelous-EN-CDF-PN-S0}$^6$ & C$^{16}$+R$^{17}$ & 5 & 3.59 & 5.38 & 37.2 & 66.3 & 43.1 & 68.1 & 98.8 & 69.4 & 57.1 & 17.5 & 59.8 \\
\textbf{Achelous-EN-GDF-PN-S0}$^7$ & C+R & 5 & 3.55 & 2.76 & 37.5 & 66.9 & 44.6 & 69.1 & 99.0 & 69.3 & 57.8 & \textbf{17.8} & \textbf{61.3} \\
\textbf{Achelous-EN-CDF-PN2-S0}$^8$ & C+R & 5 & 3.69 & 5.42 & 37.3 & 66.3 & 43.0 & 68.4 & 99.0 & 68.9 & \textbf{60.2} & 15.2 & 56.5 \\
\textbf{Achelous-EN-GDF-PN2-S0} & C+R & 5 & 3.64 & 2.84 & 37.7 & 68.1 & 45.0 & 67.2 & 99.2 & 67.3 & 59.6 & 14.8 & 57.7   \\
\textbf{Achelous-EF-GDF-PN-S0}$^9$ & C+R & 5 & 5.48 & 3.41 & 37.4 & 66.5 & 43.4 & 68.7 & \textbf{99.6} & 66.6 & 59.4 & 17.3 & 50.6  \\
\textbf{Achelous-EV-GDF-PN-S0}$^{10}$ & C+R & 5 & 3.79 & 2.89 & 38.8 & 67.3 & 42.3 & 69.8 & 99.6 & \textbf{70.6} & 58.0 & 16.4 & 54.9 \\
\textbf{Achelous-MV-GDF-PN-S0}$^{11}$ & C+R & 5 & 3.49 & 3.04 & \textbf{41.5} & \textbf{71.3} & \textbf{45.6} & \textbf{70.6} & 99.5 & 68.8 & 58.9 & 16.0 & 53.7  \\
\midrule
\textbf{Achelous-EN-GDF-PN-S1} & C+R & 5 & 5.18 & 3.66 & 41.3 & 70.8 & 45.5 & 67.4 & 99.4 & \textbf{69.3} & 58.8 & 16.6 & \textbf{59.7} \\
\textbf{Achelous-EF-GDF-PN-S1} & C+R & 5 & 8.07 & 4.52 & 40.0 & 70.2 & 43.8 & 68.2 & 99.3 & 68.7 & 58.2 & 16.6 & 46.8 \\
\textbf{Achelous-EV-GDF-PN-S1} & C+R & 5 & 4.14 & 3.16 & 41.0 & 70.7 & 45.9 & 70.1 & 99.4 & 67.9 & \textbf{59.2} & \textbf{16.7} & 56.6  \\
\textbf{Achelous-MV-GDF-PN-S1} & C+R & 5 & 4.67 & 4.29 & \textbf{43.1} & \textbf{75.8} & \textbf{47.2} & \textbf{73.2} & \textbf{99.5} & 69.2 & 59.1 & 15.8 & 55.8 \\
\midrule
\textbf{Achelous-EN-GDF-PN-S2} & C+R & 5 & 6.90 & 4.59 & 40.8 & 70.9 & 44.4 & 69.6 & 99.3 & 71.1 & \textbf{59.0} & \textbf{16.1} & \textbf{58.1} \\
\textbf{Achelous-EF-GDF-PN-S2} & C+R & 5 & 14.64 & 7.13 & 40.5 & 70.8 & 44.5 & 70.3 & 99.1 & \textbf{71.7} & 58.4 & 13.5 & 39.3  \\
\textbf{Achelous-EV-GDF-PN-S2} & C+R & 5 & 8.28 & 5.19 & 40.3 & 69.7 & 43.8 & \textbf{74.1} & 99.5 & 67.9 & 58.3 & 14.7 & 47.1 \\
\textbf{Achelous-MV-GDF-PN-S2} & C+R & 5 & 7.18 & 6.02 & \textbf{45.0} & \textbf{79.4} & \textbf{48.8} & 73.8 & \textbf{99.6} & 70.8 & 58.5 & 15.6 & 52.7 \\

\midrule

YOLOP \cite{yolop} & C & 3 & 7.90 & 18.6 & 37.9 & 68.9 & 43.5 & - & 99.0 & \textbf{74.9} & - & 1.28 & 8.15\\
HybridNets \cite{hybridnets} & C & 3 & 12.83 & 15.6 & 39.1 & \textbf{69.8} & \textbf{44.2} & - & 98.8 & 71.5 & - & 6.04 & 17.1\\
\midrule
YOLOv7-Tiny \cite{wang2022yolov7} & C &  1 & 6.03 & 33.3 & 37.3 & 65.9 & 43.7 & - & - & - & - & 36.7 & 118.6 \\
YOLOX-Tiny \cite{ge2021yolox} & C &  1 & 5.04 & 3.79 & \textbf{39.4} & 68.0 & 43.0 & - & - & - & - & 33.6 & 102.0 \\
YOLOv4-Tiny \cite{bochkovskiy2020yolov4} & C &  1 & 5.89 & 4.04 & 13.1 & 36.3 & 20.2 & - & - & - & - & \textbf{114.6} & 352.2\\
\midrule
Segformer-B0 \cite{xie2021segformer} & C &  1 & 3.71 & 5.29 & - & - & - & \textbf{72.5} & \textbf{99.2} & 72.1 & - & 41.6 & 124.7\\
PSPNet (MobileNet) \cite{zhao2017pyramid} & C &  1 & 2.38 & 2.30 & - & - & - & 69.4 & 99.0 & 69.7 & - & 61.2 & 246.1 \\
\midrule
PointNet \cite{Charles_Su_Kaichun_Guibas_2017} & R &  1 & 3.53 & 1.19 & - &- &- &- &- &- & 59.0 & 97.0 & \textbf{507.4} \\
PointNet++ \cite{Qi_Yi_Su_Guibas_2017} & R &  1 & 1.88 & 2.63 & - &- &- &- &- &- & \textbf{60.7} & 72.8 & 384.2 \\
\bottomrule
\end{tabular}
\\
\vspace{1mm}
\scriptsize{1. \textbf{TN}: task number 2. \textbf{OD}: object detection 3. \textbf{SS}: semantic segmentation 4. \textbf{WS}: waterline segmentation 5. \textbf{PC-SS}: point cloud semantic segmentation 6. EN: EdgeNeXt, CDF: CSP-Dual-FPN, PN: PointNet 7. GDF: Ghost-Dual-FPN 8. PN2: PointNet++ 9. EF: EfficientFormer 10. EV: EdgeViT 11. MV: MobileViT 12. mIoU$_{\text{t}}$: mIoU of targets 13. mIoU$_{\text{d}}$: mIoU of drivable area 14. FPS$_{\text{e}}$: FPS on Jetson AGX Xavier 15. FPS$_{\text{g}}$: FPS on RTX A4000 16. C: camera 17. R: radar.}
\end{table*}

\subsection{Experimental Settings}
 \textbf{Device.} We mount a SONY IMX-317 RGB camera and an Oculii EAGLE Imaging Radar on our USV. Sensors are temporally synchronized via timestamps and spatially synchronized via a calibration board. 

 \begin{figure}
    \centering
    \includegraphics[width=0.63\linewidth]{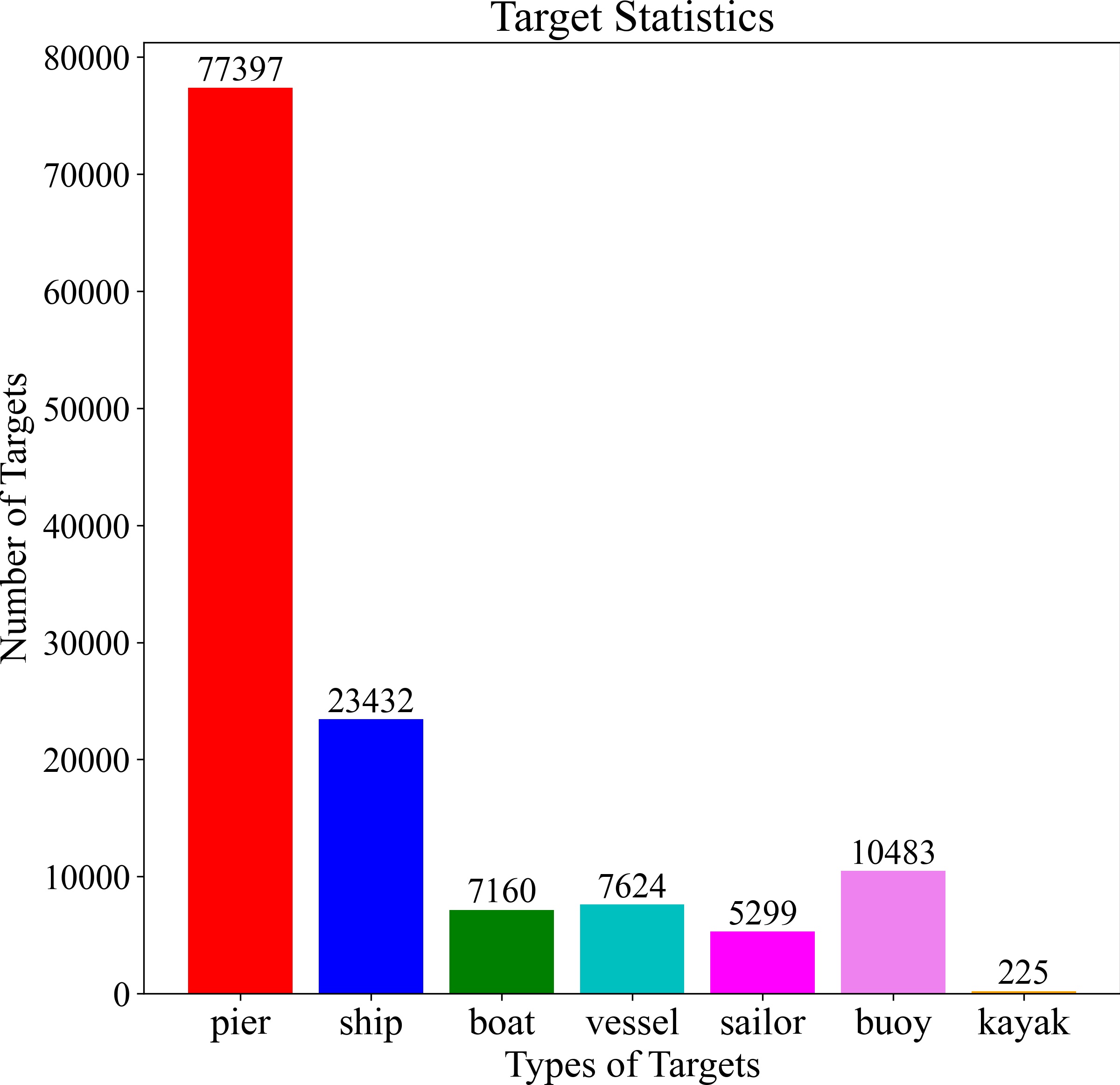}
    \caption{Statistics of our collected dataset.}
    \label{fig:dataset_statistics}
\end{figure}

 \textbf{Data.} We capture 50,000 images and frames of radar data \cite{yao2023waterscenes}. Each image is $1920 \times 1080$ pixels. There are seven classes for detection, including pier, buoy, sailor, ship, boat, vessel and kayak. Besides these classes, drivable area is an additional class for semantic segmentation while clutter is an additional class for point cloud semantic segmentation. Waterline is a single class for the waterline segmentation task. We annotate both object detection and semantic segmentation in VOC format. We annotate point cloud categories based on the ground truth of the bounding box and clustering by velocity. We divide data into the training, validation and test set by the ratio of 7:2:1.

\textbf{Training and Evaluation.} We resize images and RVP maps to $320 \times 320$ pixels. We train our Achelous for 100 epochs with a batch size of 32 and an initial learning rate of 0.03. We adopt Stochastic Gradient Descent (SGD) with a momentum of 0.937 as the optimizer and cosine learning rate scheduler. We use mixed precision and Exponential Moving Average (EMA) during training. We use the homoscedastic-uncertainty-based \cite{kendall2018multi} multi-task training strategy. We adopt focal loss in detection, dice loss in segmentation and NLL loss in point cloud segmentation. We train Achelous and other models from scratch on two RTX A4000 GPUs with data-parallel mode. We test the FPS of all models on an NVIDIA Jetson AGX Xavier (TABLE \ref{tab:jetson}) and RTX A4000. We use mAP$_{\text{50-95}}$, mAP$_{50}$ and AR$_{50}$ as metrics to evaluate object detection while mIoU is to measure semantic segmentation of both image and radar point clouds.

\begin{table}
\setlength\tabcolsep{3.5pt}
\caption{Configuration of NVIDIA Jetson AGX Xavier}
\centering
\label{tab:jetson}
\begin{tabular}{l|l}  
\toprule   
  Modules & Specifications \\
\midrule 
  Memory & 8 GB \\
  CUDA Cores & 384 NVIDIA CUDA cores + 48 Tensor cores \\
  CPU & 6-core ARM v8.2 64-bit \\
  DLA & 4.1 TFLOPS (FP16) + 8.2 TOPS (INT8)\\

\bottomrule  
\end{tabular}
\end{table}

\begin{figure*}
    \centering
    \includegraphics[width=0.85\linewidth]{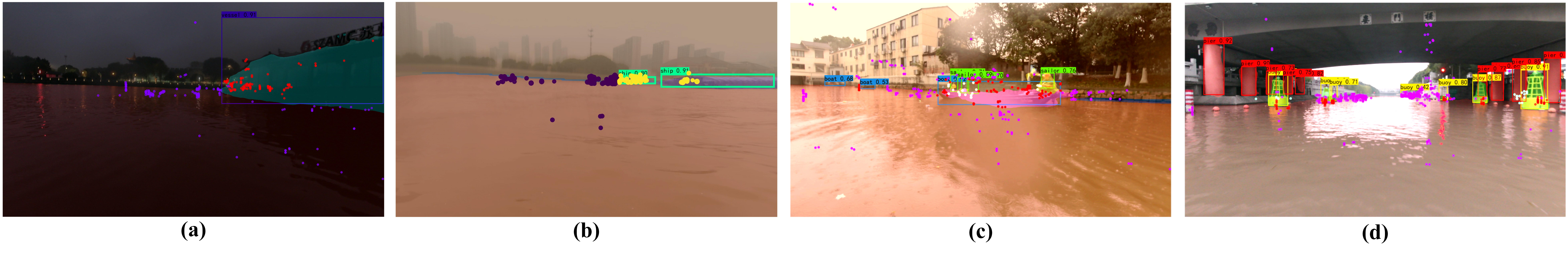}
    \caption{Panoptic Perception Results of Achelous-EN-CDF-PN-S0, object detection, segmentation of targets, drivable area and waterline, and point cloud semantic segmentation. (a) Dark environment. (b) Occluded ships on a dense foggy day. (c) The lens blocked by water droplets. (d) Dense targets.}
    \label{fig:prediction_results}
\end{figure*}

\begin{figure*}
    \centering
    \includegraphics[width=0.85\linewidth]{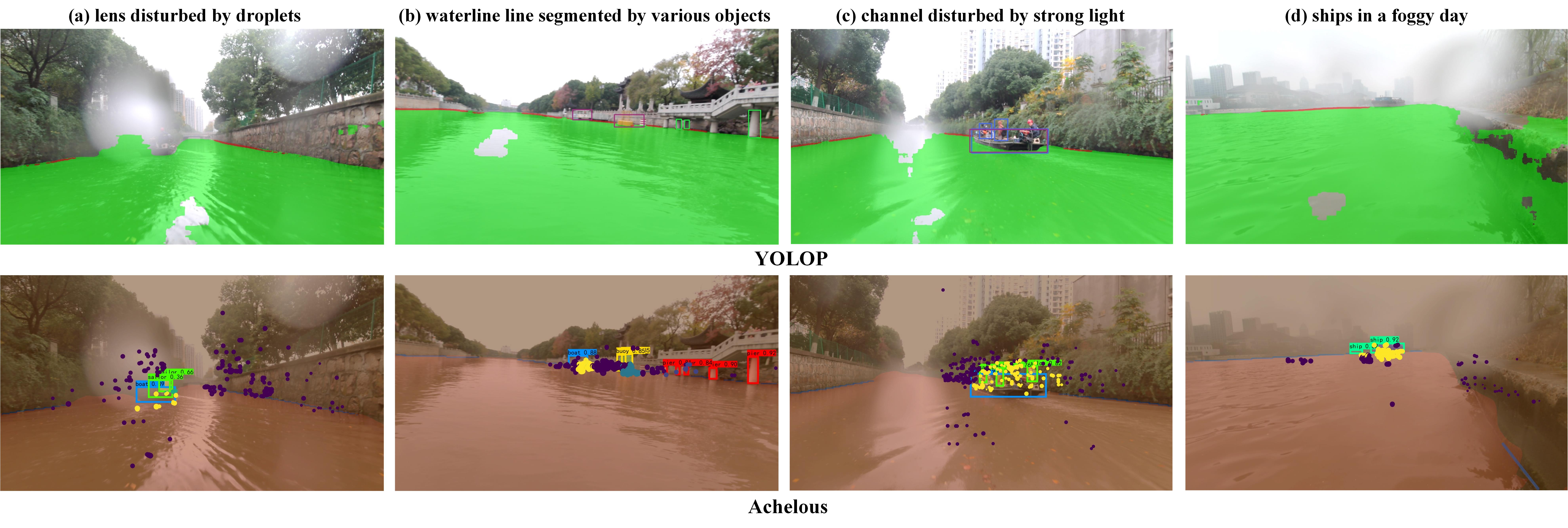}
    \caption{Comparison of Achelous-MV-GDF-PN-S0 (4.4 million parameters less than YOLOP) with YOLOP under various situations.}
    \label{fig:yolop_compare}
\end{figure*}

\begin{figure}
    \centering
    \includegraphics[width=0.98\linewidth]{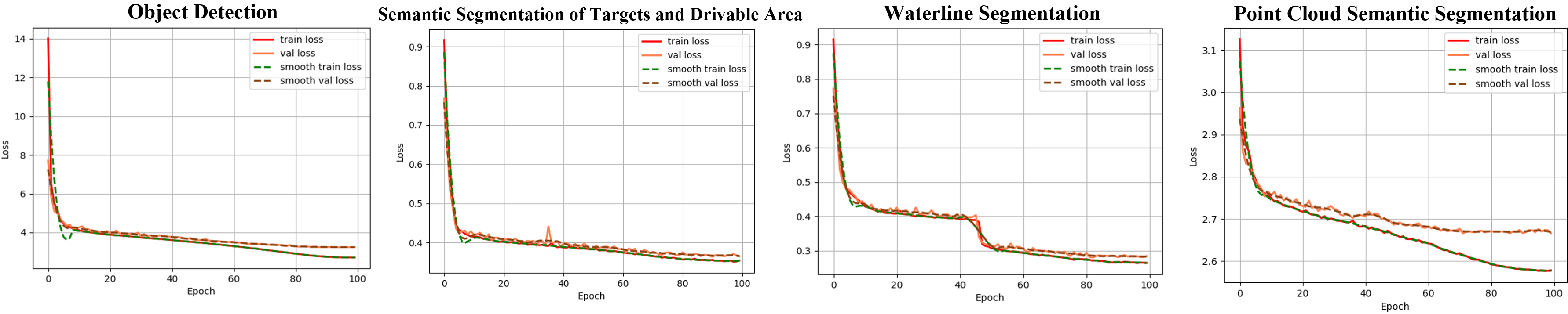}
    \caption{Training and Validation Loss of Achelous-EN-GDF-PN2-S0.}
    \label{fig:loss_example}
\end{figure}

\subsection{Comparison of Achelous with Other Models}
We compare our Achelous with other panoptic perception models and single-task models. We evaluate performances of object detection, target semantic segmentation, drivable area segmentation, waterline segmentation and point cloud semantic segmentation. We also test FPS on an edge device (NVIDIA Jetson AGX Xavier) and a high-performance GPU (RTX A4000). We observe that Achelous converges normally during multi-task training (Fig. \ref{fig:loss_example}). As Table \ref{tab:prediction_results} presents, our Achelous achieves state-of-the-art performances on object detection, semantic segmentation of objects and drivable area compared with other panoptic perception models and single-task models. We observe that Achelous with the backbone of MobileViT achieves the best performance on object detection and semantic segmentation for three sizes (S0, S1 and S2), considerably outperforming other models. However, for waterline segmentation, YOLOP outperforms Achelous by about 3\% mIoU. For point cloud semantic segmentation, Achelous with PointNet++ outperforms Achelous with PointNet. Furthermore, Achelous is much faster than YOLOP and HybridNets. The FPS of Achelous is between 13 to 18 on an NVIDIA Jetson AGX Xavier, which satisfies real-time inference for autonomous driving of USVs at a high speed.

We also visualize the prediction results of Achelous and YOLOP in Fig. \ref{fig:prediction_results} and Fig. \ref{fig:yolop_compare}. We can see in most circumstances, our Achelous can better detect and segment targets than YOLOP, no matter in dark environments, adverse weather or light interference.

\begin{table}
\setlength\tabcolsep{2.3pt}
\caption{Inference Speed of Standalone Models and Achelous}
\centering
\label{tab:speed_compare}
\begin{tabular}{l|c|c|c}  
\toprule   
  \textbf{Methods} & \textbf{Tasks} & \textbf{Latency$_{\text{g}} ^1$ (ms)} & \textbf{Latency$_{\text{e}} ^2$ (ms)}  \\

\midrule
  YOLOX-Tiny \cite{ge2021yolox} & OD  & \multirow{4}[2]{*}{21.2}  & \multirow{4}[2]{*}{79.5}  \\
  Segformer-B0 \cite{xie2021segformer} & SS &  \\
  PSPNet \cite{zhao2017pyramid} & WS &   \\
  PointNet \cite{Charles_Su_Kaichun_Guibas_2017} & PC-SS &  \\
\midrule 
  Achelous & OD \& SS  & \multirow{2}[2]{*}{16.3 ($\downarrow 4.9$)}  &  \multirow{2}[2]{*}{56.2  ($\downarrow 23.3$)}   \\
     (EN-GDF-PN-S0)                                    &  WS \& PS-SS &      \\
  
\bottomrule  
\end{tabular}
\\
\vspace{1mm}
\footnotesize{1. \textbf{Latency$_{\text{g}}$}: latency on a RTX A4000 GPU 2. \textbf{Latency$_{\text{e}}$}: latency on an NVIDIA Jetson AGX Xavier.}\\
\end{table}

\subsection{Speed of Parallel Standalone Models and Achelous}
As TABLE \ref{tab:speed_compare} presents, we test the latency of standalone models of parallel inference and our Achelous-EN-GDF-PN-S0 on both Jetson AGX Xavier and RTX A4000. Our Achelous's latency is lower than parallel standalone models when conducting several panoptic perception tasks, no matter on Jetson or RTX A4000. It proves that multi-task models are necessary for panoptic perception to improve efficiency.

\begin{figure*}
    \centering
    \includegraphics[width=0.87\linewidth]{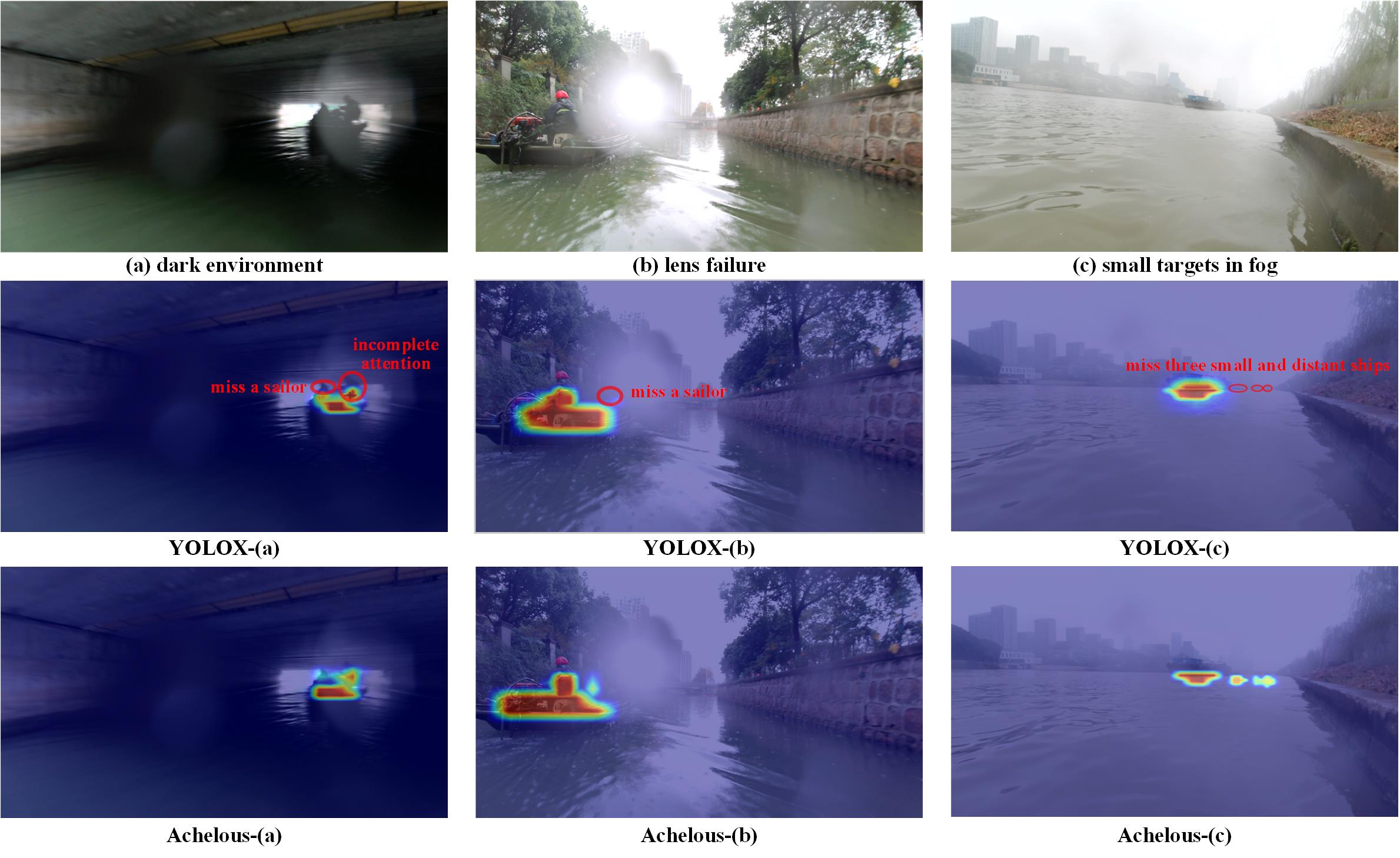}
    \caption{Visualization of heatmaps of Achelous-MV-GDF-PN-S2 (7.2M parameters) and YOLOX-M (25.3M parameters) in different situations.}
    \label{fig:heatmap_visual}
\end{figure*}

\subsection{Ablation Experiments}

\begin{table}
\caption{Ablation Experiment of RCNet and Fusion Methods}
\centering
\label{tab:rc_net_compare}
\begin{tabular}{l|ll}  
\toprule   
  \textbf{Methods} & \textbf{mAP}$_{\text{50-95}}$ & \textbf{mAP}$_{\text{50}}$  \\
\midrule 
  Achelous-MobileNetv2 & 37.2  & 66.0  \\
  Achelous-RCNet & 37.5 ($\uparrow 0.3$) & 66.6 ($\uparrow 0.6$)   \\
\midrule
  \textbf{Fusion Methods} & \textbf{mAP}$_{\text{50-95}}$ & \textbf{Latency$_{\text{e}}$ (ms)} \\
\midrule
  Backbone + Dual-FPN & 37.4 $\pm 0.3$ & 29.5 \\
  Dual-FPN & 37.5 $\pm 0.2$ & 16.3 ($\downarrow 13.2$) \\
\bottomrule  
\end{tabular}
\end{table}

We first do the ablation experiment on RCNet (TABLE \ref{tab:rc_net_compare}), where we replace RCNet with MobileNetV2 in Achelous-EN-GDF-PN-S0, a structurally similar network consisting of normal convolutions. We notice that mAP$_{\text{50-95}}$ drops $0.2$ while AP$_{\text{50}}$ drops about 0.5. It proves that RCNet containing Radar Convolution can better capture and model features of radar point clouds than the normal convolution calculator.

Furthermore, we compare the results of two different fusion methods. We find that fusion of image and radar features in both the backbone and fpn stage could not improve detection performances notably, whose inference latency is 13.2 ms slower than the fpn-level fusion.

\subsection{Visualization and Analysis of Feature Maps}

To validate whether our Achelous pays attention to correct regions of interest, we adopt Grad-CAM \cite{selvaraju2017grad} to visualize the heatmaps of the last layer in FPN, which is connected with the detection head. We choose YOLOX-M to compare it with our Achelous-MV-GDF-PN-S2. We choose three challenging scenarios: a dark environment under the bridge, lens failure by droplet and a foggy day. Firstly, we observe that vision-only YOLOX-M performs terribly in the dark environment, where a sailor is missed and another is not focused preciously by YOLOX-M. Excitingly, Achelous with vision and radar features captures the distant sailor successfully. Secondly, when confronted with droplets on the lens, vision-only YOLOX-M completely ignore the targets in the region disturbed by droplets, but Achelous notices the ignored sailor. Thirdly, for the circumstance that three distant ships are obscured by the thick fog, our Achelous based on vision-radar fusion is aware of three small and distant ships but YOLOX-M is not, which validates the radar with long-range detection capability matters in some adverse weather. In all, Achelous based on the feature-level fusion of camera and 4D radar with fewer parameters is much more reliable than vision-only models during various challenging situations.

\section{Conclusion}
We propose a powerful and scalable riverway panoptic perception framework called Achelous based on camera and 4D mmWave radar, which can simultaneously perform five different perception tasks of vision-level and point-cloud-level. Achelous is a high-efficiency framework, inferring in real-time on an NVIDIA Jetson AGX Xavier. We also propose radar convolution, which can exquisitely extract sparse and irregular features of radar point clouds. Achelous also outperforms other panoptic perception models and single-task models on most perception tasks, especially in adverse situations. We hope Achelous can promote the development of water-surface panoptic perception, providing a low-cost and high-efficiency scheme for researchers.





\bibliographystyle{IEEEbib}
\bibliography{root}

\end{document}